\documentclass{Interspeech2024}

\usepackage{adjustbox}
\usepackage{multirow}
\usepackage{colortbl}
\usepackage{subcaption}
\usepackage{cleveref}
\usepackage{lipsum}

\usepackage[acronym, shortcuts, nohypertypes={acronym}]{glossaries}
\newacronym{LOSO}{LOSO}{Leave-One-Speaker-Out}
\newacronym{CV}{CV}{cross-validation}
\newacronym{CI}{CI}{confidence interval}
\newacronym{SVM}{SVM}{Support Vector Machine}
\newacronym{ML}{ML}{Machine Learning}
\newacronym{LLD}{LLD}{low-level descriptor}
\newacronym{w2v}{w2v2}{wav2vec2.0}
\newacronym{eGeMAPS}{eGeMAPS}{extended Geneva minimalistc acoustic parameter set}

\newcommand{\eg}{e.\,g.}

\interspeechcameraready

\usepackage[natbib, bibencoding=utf8, citestyle=numeric, bibstyle=ieee, minbibnames=6, maxbibnames=10, maxcitenames=2, mincitenames=1, sortcites]{biblatex}
\bibliography{mybib}

\title{Exploring Gender-Specific Speech Patterns in Automatic\\Suicide Risk Assessment}

\name[affiliation={1}]{Maurice}{Gerczuk}
\name[affiliation={2}]{Shahin}{Amiriparian}
\name[affiliation={3}]{Justina}{Lutz}
\name[affiliation={3}]{Wolfgang}{Strube}
\name[affiliation={3}]{Irina}{Papazova}
\name[affiliation={3,4}]{Alkomiet}{Hasan}
\name[affiliation={1,2,5}]{Björn W.}{Schuller}

\address{
  $^1$Chair of Embedded Intelligence for Health Care \& Wellbeing, University of Augsburg, Germany \\
  $^2$CHI -- Chair of Health Informatics, MRI, TU Munich, Germany \\
  $^3$District Hospital Augsburg, Germany \\
  $^4$German Center for Mental Health, Munich, Germany \\
  $^5$GLAM -- Group on Language, Audio, \& Music, Imperial College, UK
}
\email{maurice.gerczuk@uni-a.de}

\keywords{suicidality, computational paralinguistics, digital health}

\usepackage{xcolor}

\begin{document}

\maketitle

\begin{abstract}
    
In emergency medicine, timely intervention for patients at risk of suicide is often hindered by delayed access to specialised psychiatric care. To bridge this gap, we introduce a speech-based approach for automatic suicide risk assessment. Our study involves a novel dataset comprising speech recordings of $20$ patients who read neutral texts. We extract four speech representations encompassing interpretable and deep features. Further, we explore the impact of gender-based modelling and phrase-level normalisation. By applying gender-exclusive modelling, features extracted from an emotion fine-tuned wav2vec2.0 model can be utilised to discriminate high- from low suicide risk with a balanced accuracy of $81\,\%$. Finally, our analysis reveals a discrepancy in the relationship of speech characteristics and suicide risk between female and male subjects. For men in our dataset, suicide risk increases together with agitation while voice characteristics of female subjects point the other way.
\end{abstract}


\section{Introduction}
Every year, more than 700\,000 people die by suicide, accounting for 1 in every 100 deaths. For young people, suicide is the fourth leading cause of death following road injury, tuberculosis, and interpersonal violence~\cite{world2021suicide}. Main factors contributing to suicidality can be economic pressure, past mental health issues, and personal or external crises, such as the COVID-19 pandemic which lead to an increase in suicidal ideation~\cite{Yan23-SBA}.

There are gender-specific differences in the frequency of suicidal behaviour and suicides. Women are more likely to exhibit suicidal behaviour, while suicide is more common in men. This phenomenon, known as the gender paradox~\cite{Canetto98-TGP}, is in part explained by the different preferences of men and women regarding the methods of suicide they choose. Overall, men choose more lethal suicide methods like firearms or hanging~\cite{Callanan12-GDI}, also referred to as ``violent'' suicide methods~\cite{Ludwig18-TCO}. However, it also appears that the rate of completion among men is higher even when using the same suicide methods as women~\cite{Cibis12-POL}. ~\citet{Freeman17-ACS} found that the intent to die when committing a suicide attempt was greater among men than women resulting in more serious suicide attempts vs
suicide attempts that are being made as a cry for help or a means of self-harm. Evaluating data from 78 countries over 11 years, ~\citet{Milner20-SIG} showed that whilst rising gender equality led to reduced suicide rates in women, there was no significant reduction in suicide rates amongst males. The assumption of several roles in an equal social system was seen as having positive effects on female mental health whilst masculine norms linked with increased vulnerability for suicidal ideation are not being overcome at the same pace. 

As professional psychiatric assessment of suicide risk is costly and often not available in emergency medicine, predictors which are easy and quick to collect are highly valuable~\cite{Iyer22-DOS}. However, a meta-study showed that traditional analysis of individual risk factors through questionnaires is of limited accuracy when predicting elevated suicide-risk~\cite{Franklin17-RFF}. In this context, biomarkers can provide a promising source of information, complementing the standard practice of screening for suicide attempt history~\cite{Sudol17-BOS}. As a biosignal that is both easy to collect as well as connected to both the physiological and cognitive systems, speech can be harnessed to analyse a wide range of health conditions~\cite{Cummins18-SAF}. For depression and suicidality, both traditional statistical analysis and machine learning strategies can be utilised to automatically infer aspects of a person's current mental health from various linguistic and paralinguistic characteristics ~\cite{Cummins15-ARO}.

\citet{Min23-AAO} fused voice characteristics with demographic information and the number of suicide attempts to classify high- against low-suicide risk and worsening of suicidality with accuracies of $69\,\%$ and $79\,\%$, respectively.  \citet{Belouali21-AAL} utilised acoustic and linguistic features of speech recorded from army veterans in order to automatically assess suicidal ideation using \ac{ML}. Based on an acoustic feature analysis, they found the voices of suicidal subjects in their exclusively male cohort to be more monotonous, dull and breathy, a finding in line with voice characteristics of suicidal adolescents~\cite{Scherer13-ITS}. Further, a person's speech fluency has been found to be a discriminative feature, as it is negatively impacted by suicidal ideation and suicide attempts ~\cite{Stasak21-RSV}.

Despite the differences in suicidal behaviour between women and men, few research directly targets an investigation into gender-specific correlates of speech characteristics and suicide risk. However, for the related subject of depression recognition from speech, ~\citet{Vlasenko17-IGV} found there to be a significant difference in the manifestation of depression in the formants of vowels between genders, leading to improved classification performance when utilising gender-specific features. Furthermore, \citet{Oureshi21-GEO} showed that adversarially learning to predict depression scores separated by gender leads to a more accurate estimation of depression severity on the DAIC-WOZ corpus.

In the presented work, we want to extend previous research on automatic suicide risk assessment from speech by (1) applying state-of-the-art transformer-based speech representations and comparing their efficacy against traditional audio functionals; and (2) exploring differences in speech patterns between female and male subjects via gender-based modelling and manual acoustic analysis.

\section{Dataset}

Our database includes speech recorded from 10 women (age $18-61$\,years, $\mu=41.7 \pm 15.5$) and 10 men (age $18-64$\,years, $\mu=37.7 \pm 16.3$) undergoing emergency admission to the psychiatric department of the District Hospital Augsburg, Germany\footnote{Ethical approval for the study was obtained from the ethics commission of the University of Augsburg.}. Due to the setting and targeted application, the dataset does not include samples recorded from healthy controls, rather, every subject is diagnosed with at least one behavioural or mental disorder according to ICD-10. A majority ($N=15$) were experiencing a severe depressive episode without psychotic symptoms which mostly occurred as part of recurrent depressive disorder (F33.2, $N=13$). Other main diagnoses were paranoid schizophrenia (F20.0, $N=1$), acute polymorphic psychotic disorder with ($F23.1, N=1$) and without symptoms of schizophrenia ($F23.0, N=1$), post-traumatic stress disorder (F43.1, $N=1$), and adjustment disorder (F43.2, $N=1$). The study doctor assessed the near-term suicide risk of each participant on a Likert scale of [1 -- 6] (1--2: no risk, 3--4: suicidal without intention to act, 5--6: high suicide risk with intention to act). In this paper, we target a binary classification of high, near-term suicide risk (5--6) against lower suicidality (1--4). Applying this cutoff, 7 subjects -- 4 male and 3 female -- were at high risk of suicide.
Three types of speech were obtained from every subject using a Zoom Q8 recorder: (1) two repetitions of readings of neutral texts\footnote{Three German short stories ``Der Nordwind und die Sonne'', ``Gleich am Walde'', and ``Der Hund und das Stück Fleisch''.}; (2) a spontaneous description of a comic; and (3) isolated vowel production. In this study, we focus our analysis on the neutral texts.

\section{Methodology}
Our approach begins with a preprocessing step where we normalise the volume of the audio samples and segment them into phrases via forced alignment. Afterwards, we extract both interpretable audio functionals and deep features from the segments and apply either global or phrase-level normalisation. We train and evaluate classifiers in a \ac{LOSO} \ac{CV} and investigate the effects of gender-based modelling.

\subsection{Segmentation}
After applying loudness normalisation, we break down the recordings of neutral texts into individual sentences. We apply forced alignment of the texts to the audio signals via whisperx~\cite{Bain23-WTS} and split the recordings at the detected sentence boundaries. After segmentation, we receive 6, 7, and 16 segments for each recording of the three stories respectively, resulting in $1\,160$ speech samples across two readings of all three stories by every participant with a mean duration of $5.06$\,s ($\pm 3.29$\,s).

\subsection{Feature extraction}
After segmentation, we extract audio representations from each phrase separately. We follow two general paradigms of paralinguistic speech analysis by computing both interpretable sets of audio functionals as well as generating deep feature embeddings from pre-trained large-scale speech recognition models based on \ac{w2v}.

\noindent
\textbf{Audio Functionals:}
We extract two sets of interpretable audio features in the form of openSMILE~\cite{Eyben10-OTM} functionals -- the handcrafted \ac{eGeMAPS}~\cite{Eyben16-TGM} set ($N=88$) and the larger \textsc{ComParE2016} ($N=6,\,373$) feature set. Both include a range of statistical functions (e.\,g., mean, minimum, maximum, standard deviation, etc.) applied to \acp{LLD} computed over consecutive short frames of the audio signal. The \acp{LLD} include parameters related to frequency (e.\,g., F0, jitter, centres and bandwidths of formats 1-3), energy (e.\,g., loudness, shimmer and harmonic to noise ratio), spectral balance and shape (e.\,g., Alpha Ratio, Hammarberg Index, MFCCs and spectral slopes), and temporal features (e.\,g., durations and frequencies of voiced segments or loudness peaks).

\noindent
\textbf{\ac{w2v} embeddings:}
We further utilise the mean of the hidden states of the last encoder layer in pre-trained \ac{w2v} models. We specifically select two pre-trained transformer models: wav2vec-large~\cite{Baevski20-W2A} and a 12-layer \ac{w2v}  model fine-tuned for dimensional speech emotion recognition (arousal, dominance, and valence) on MSP-Podcast~\cite{Lotfian19-BNE,Wagner23-DOT}. The former allows us to analyse whether features learnt in a self-supervised manner large-scale multi-lingual speech data capture paralinguistic markers of suicidality while the latter might be able to exploit correlations between a subject's emotional state and suicide risk.

\subsection{Feature Normalisation}

We apply conventional \emph{global} feature normalisation by scaling every feature to zero mean and unit variance across the respective training datasets. Furthermore, we exploit the fixed-content nature of the collected speech samples by scaling the features on the \emph{phrase} level. Specifically, we define the phrases as the individual sentences in the three neutral texts. We scale each sample according to statistics computed across all samples of the same sentence. 

\subsection{Suicide Risk Classification}
All experiments utilise \acp{SVM} with linear kernel to classify individual phrases (either sentences or vowels). We tune 
the \ac{SVM}'s cost parameter on a logarithmic scale between $10^{-2}$ and $10^{-7}$ and balance the weights of high and low suicide risk samples during training based on their frequency. 

\noindent
\textbf{Gender-Based Modelling: }
We investigate how gender-related differences in voice characteristics affect suicidality recognition performance. For this purpose, we evaluate two strategies for building gender-specific classification systems. In the first case, each model's parameters and normalisation statistics are learnt exclusively on samples of the respective gender, effectively partitioning the data before applying the pipeline. Secondly, we experiment with ``soft'' gender-based modelling by using weighted instance learning. Here, instead of excluding all samples of the other gender(s) from each model's training data, we apply a down-weighting factor $\lambda=0.1$ to every out-of-group instance.

\noindent
\textbf{Evaluation Strategy: }
Due to the small size of our database, we utilise \ac{LOSO} \ac{CV}. Normalisation statistics are computed on each fold's training data and hyperparameters are optimised via nested \ac{CV} (5 inner folds). For model evaluation and hyperparameter tuning, we choose balanced accuracy -- the mean of per class recalls -- as a metric accounting for class imbalances. As the models are trained to classify suicidality per phrase but ground truth labels were assigned per subject, we present both segment-level and majority-voted subject metrics. Additionally, we report $95\,\%$ \acp{CI} computed over $1\,000$  bootstrapped samples of the model predictions. Note that bootstrapping is applied before majority voting, sometimes resulting in the actual subject level metric lying at the edge of the \ac{CI}. 

\section{Experiments and Results}
In the following, we first discuss the results of our machine learning experiments for different sets of audio features, normalisation strategies and gender-based modelling. Afterwards, we perform a feature analysis to identify the most important acoustic markers for high suicide risk.

\subsection{Classification Results}
\label{sec:results}
\begin{table*}[h]
    \centering
\resizebox{\textwidth}{!}{
\begin{tabular}{ll|cc|cc|cc}
\toprule
\textbf{Balanced Accuracy [\%]} & & \multicolumn{2}{c|}{\textbf{Global Modelling}} & \multicolumn{4}{c}{\textbf{Gender-based Modelling}} \\
 & & \multicolumn{2}{c|}{} & \multicolumn{2}{c|}{\textbf{$\lambda=0$}} & \multicolumn{2}{c}{$\lambda=0.1$} \\
\textbf{Features} & & global normalisation & phrase normalisation & global normalisation & phrase normalisation & global normalisation & phrase normalisation \\
\midrule
\multirow[t]{2}{*}{ComParE2016} & & 52 (49-55) / 59 (45-63) & 55 (52-57) / 59 (48-66) & 62 (59-65) / 63 (55-70) & 61 (59-64) / 59 (59-70) & 61 (57-63) / 70 (63-78) & 64 (61-67) / 70 (66-78) \\
\multirow[t]{2}{*}{eGeMAPSv02} & & 58 (55-61) / 63 (49-70) & 54 (51-57) / 46 (38-53) & 56 (53-59) / 60 (60-64) & 52 (49-55) / 60 (46-67) & 60 (57-63) / 60 (60-64) & 54 (51-57) / 49 (46-57) \\
\midrule
\multirow[t]{2}{*}{wav2vec2-audeering-emo-dim} & & 57 (54-60) / 49 (45-56) & 54 (51-56) / 45 (41-52) & 66 (63-69) / 74 (67-82) & \cellcolor{gray!40}\textbf{70 (68-73) / 81 (70-85)} & 54 (52-57) / 49 (38-56) & 58 (55-61) / 52 (48-60) \\
\multirow[t]{2}{*}{wav2vec2-large-xlsr-53} & & 52 (49-55) / 53 (49-60) & 55 (52-58) / 65 (52-69) & 60 (57-63) / 56 (49-67) & 59 (56-63) / 67 (56-74) & 58 (55-61) / 67 (56-71) & 61 (58-64) / 74 (60-74) \\
\bottomrule
\end{tabular}
}
    \caption{Classification results for audio-based suicide risk assessment measured in balanced accuracy computed sample-wise and additionally aggregated per speaker by majority vote (after forward slash). $95\,\%$ confidence intervals computed from bootstrapped model predictions are given in parentheses. Both global -- one classifier is trained to generate predictions for all samples -- and gender-based modelling is evaluated. For gender-based modelling, each gender's classifier is trained via weighted instance learning, applying a down-weighting factor $\lambda$ to all samples of the other gender(s). In our dataset, only women and men are represented.}
    \label{tab:results}
\end{table*} 

\Cref{tab:results} shows the results achieved for binary suicide risk classification utilising the four evaluated audio features and ~\acp{SVM} measured in balanced accuracy. Furthermore, global modelling is compared against two variations of gender-based modelling and features are normalised either globally or per phrase. Generally, we observe a positive impact of gender-based modelling on classification accuracy for every set of audio features, except \ac{eGeMAPS} which further performs the weakest overall, only reaching a maximum $60\,\%$ segment-level and $63\,\%$ speaker-level balanced accuracy. The benefits of training gender-specific models become apparent in the larger, brute-force ComParE2016 feature set. While by itself, the models trained on these features rank below \ac{eGeMAPS}, possibly due to overfitting, separating by gender helps reduce the feature space size, minimising the impact of acoustic information unrelated to suicide risk. Moreover, down-weighting ($\lambda = 0.1$) the opposite gender samples instead of completely removing them from the training data ($\lambda = 0$) increases per speaker accuracies up to $70\,\%$ while sample-level performance stays the same, indicating robustness towards inter-subject variations and increased generalisation capabilities. Finally, neither \ac{eGeMAPS} nor ComParE2016 benefit from applying input normalisation per phrase instead of globally, indicating that models can easily abstract from the influence of linguistic content on these feature sets. This is further in line with findings for speech emotion recognition, where phonetic variations only have a small impact on model performance when utilising spectral features~\cite{Sethu09-SDO}.

However, the results achieved with models trained on \ac{w2v} embeddings paint a slightly different picture. Here, likely due to overfitting, baseline performance without either phrase normalisation or gender-based modelling is quite poor for features extracted from both the large cross-lingual and the emotion finetuned \ac{w2v}. In the case of the large XLSR model, applying phrase normalisation consistently improves per-speaker accuracy. As pointed out by ~\citet{Wagner23-DOT}, \ac{w2v} models implicitly capture linguistic information, which, however, is irrelevant to the task at hand, considering the fixed content of the neutral texts we base our analysis on. Phrase-level normalisation might help remove this confounding factor at the beginning of the machine learning pipeline. Gender-based modelling further helps performance, with the soft, down-weighting variant ($\lambda = 0.1$) slightly edging out gender-exclusive models with a best balanced accuracy of $74\,\%$. 

\begin{figure}
    \centering
    \includegraphics[width=\columnwidth]{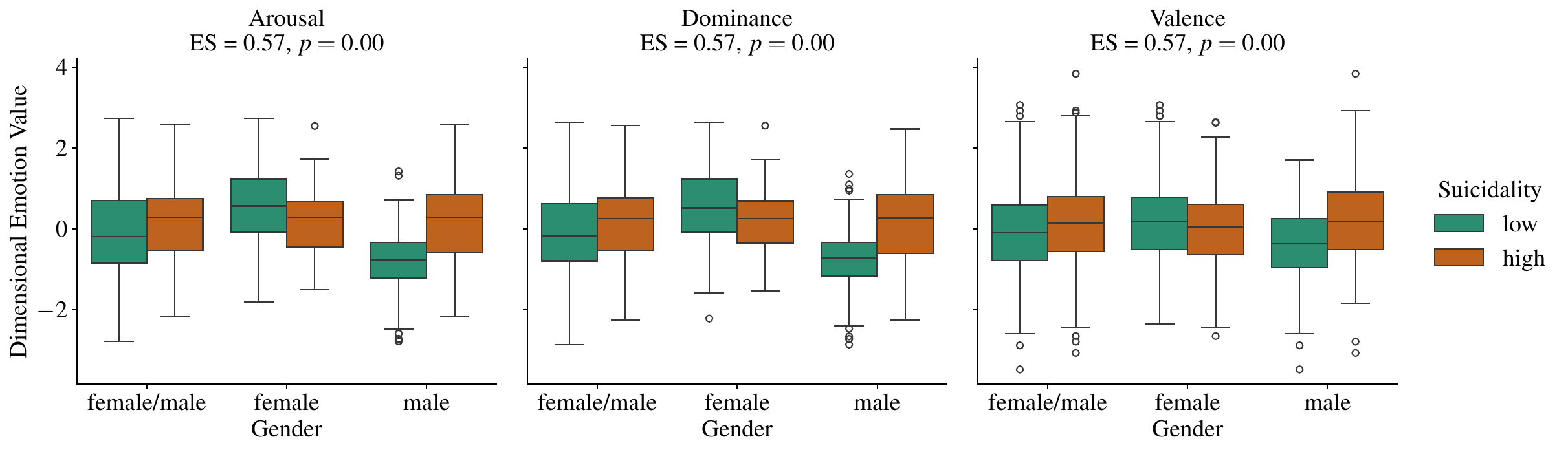}
    \caption{Distribution of normalised arousal, dominance, and valence predictions generated by the pre-trained \ac{w2v} speech emotion recognition model~\cite{Wagner23-DOT} in subjects with high and low suicidal risk additionally separated by gender.}
    \label{fig:emo-boxes}
\end{figure}

Interestingly, the best overall classification accuracy is achieved with the emotion-finetuned \ac{w2v} features, but only under gender-exclusive modelling. The fact that neither phrase normalisation nor soft gender-based modelling can help models exceed near chance-level performance, could hint towards a relationship between affective states and suicidality that is specific to gender. In order to investigate this hypothesis, we extract dimensional arousal, dominance, and valence scores for every audio sample via the same fine-tuned \ac{w2v} model used to generate the feature embeddings. We visualise the distributions of the normalised values in ~\Cref{fig:emo-boxes}. For high-risk male subjects, all three dimensions, but especially arousal, are distinctly higher than for low-risk men. Contrarily, this trend seems reversed when we look at the female subjects where higher values in the dimensions are associated with lower suicide risk. In line with these observations, ~\citet{Bryan14-GDI} found that increased agitation (a high arousal affective state) was significantly associated with the history of attempted suicide in men but did not impact suicidality for women in the same study. 

\subsection{Acoustic Feature Analysis}
\begin{figure*}
\centering
    \includegraphics[width=\textwidth]{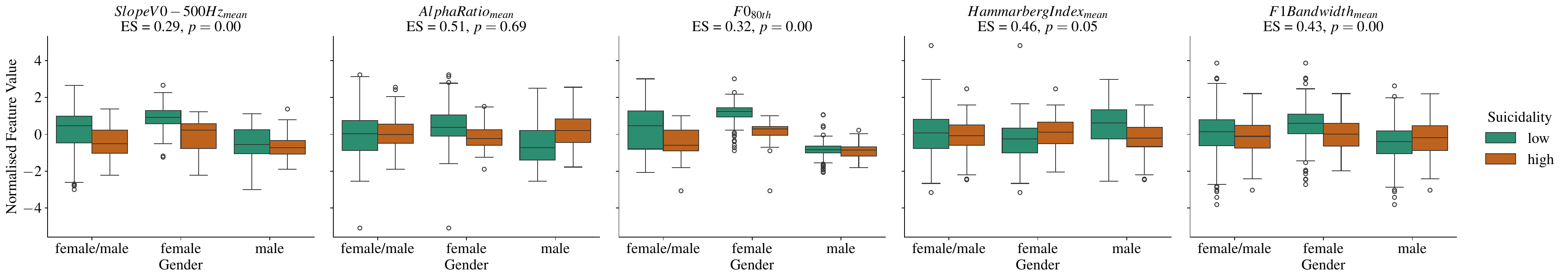}
    \caption{Distribution of most important features determined by effect size of Mann-Whitney U test for low and high suicidality. Additionally split by gender of speaker.}
    \label{fig:feature-boxes}
\end{figure*}

To build on these observations, we now investigate the relationship between paralinguistic speech characteristics and suicide risk -- and how this relation differs across the two genders present in our dataset -- by analysing a selection of acoustic features. In order to determine the most relevant features, we rank the functionals contained in \ac{eGeMAPS} according to the mean absolute coefficients (feature weights) assigned to them during the training of the linear \ac{SVM} models (configuration in the first column of \Cref{tab:results}). After ranking, we choose the top 5 functionals, sorting out redundant features, e.\,g., we remove the arithmetic mean of F0 as it is highly related to the 80th percentile of F0 which has a higher ranking. For each feature, we compute the non-parametric two-sided Mann-Whitney U test~\cite{Mann47-OAT} comparing the globally normalised feature distributions between low- and high-risk subjects. These distributions, additionally disaggregated by gender, and the corresponding common language effect sizes -- the proportion of low and high-risk feature pairs where the value of the high-risk sample is greater than that of the low-risk sample -- and p-values are visualised in~\Cref{fig:feature-boxes}.

\noindent
$\mathbf{SlopeV0-500\,Hz_{mean}}$:
We start with the linear regression slope fitted to the logarithmic power spectrum of voiced frames in the 0-500\,Hz band  which shows a negative correlation with suicide risk for women but only a negligible connection to suicidality for male subjects. The spectral slope has previously been shown to decrease with negative, low arousal emotions~\cite{Guzman13-IOS} and is further negatively affected by depression~\cite{Cummins11-AIO,Honig14-AMO}. Moreover, ~\citet{Ozdas04-IOV} found that the spectral slope decreases between healthy controls and both depressed and near-term suicidal individuals, but patients with higher suicide risk exhibit increased glottal spectral slopes compared to depressed, non-suicidal subjects. However, their analysis was only based on male subjects.

\noindent
$\mathbf{Alpha Ratio_{mean}}$:
Analysed across the whole dataset, there is no relationship between the ratio of spectral energy above and below $1\,000$\,Hz and suicidality. However, when looking at the genders separately, a shift of spectral energy towards 1\,kHz -- 5\,kHz (higher alpha ratio) is associated with an increased suicide risk for male subjects while the opposite relationship can be observed for women. Higher alpha ratio can be a product of increased emotional arousal or activity~\cite{Waaramaa10-POE}, aligning with the behaviour of the emotion fine-tuned \ac{w2v} outlined in \Cref{sec:results}

\noindent
$\mathbf{HammarbergIndex_{mean}}$:
Defined as the difference between the maximum energy peak in the 0-2\,kHz and the 2-5\,kHz bands, the Hammarberg Index is considered an indicator of vocal effort~\cite{Hammarberg80-PAA}. Like the alpha ratio, an energy increase in the upper-frequency bands (lower Hammarberg Index) is  often associated with higher arousal and basic emotions such as hot anger~\cite{Banse96-API}. Accordingly, we observe a decrease in the Index with higher suicide risk in male subjects, while a less pronounced trend in the opposite direction can be found for women.

\noindent
$\mathbf{F0_{80th}}$:
While no difference in mean F0 between low- and high-risk men exists in our dataset, women in the high-risk group display significantly reduced pitch, a feature that is often correlated with depression severity~\cite{Cummins15-ARO}. 

\noindent
$\mathbf{F1Bandwidth_{mean}}$:
Finally, the bandwidth of the first formant narrows with higher suicide risk for female subjects, indicating less breathiness. Contrary to the trends observed in the other analysed features, this is somewhat atypical for speech produced under increased depressive symptom severity~\cite{Cummins22-MMO}.

Overall, the acoustic feature analysis further supports our findings with the emotion fine-tuned \ac{w2v} model. Specifically, suicidality is reflected differently in the voices of women and men in our dataset. While high-risk female subjects exhibit speech patterns that align with an increase in depressive symptomatology (\eg, decreased pitch and narrowed spread of spectral energy), higher suicide risk in men comes with paralinguistic characteristics that are associated with heightened activation and emotional arousal.
\section{Conclusions and Future Work}
In the present work, we utilised a novel database for the automatic assessment of suicide risk from speech in emergency medicine and investigated gender-specific correlations between suicidality and paralinguistics. Our machine learning results and a consecutive acoustic feature analysis indicate that gender-based modelling can be effective for discriminating between low- and high-risk individuals. In our data, high suicidality in men positively correlates with paralinguistic characteristics that are usually associated with a high-arousal, agitated affective state while the speech patterns of female subjects point in the opposite direction. This finding could be connected to existing literature on suicide risk~\cite{Bryan14-GDI} which shows a male-exclusive correlation of agitation and suicide attempt history. 
The results further indicate the efficacy of deep, transformer-based speech representations, even in a small database. A \textbf{limitation} of our study can be found with the small size of our database, containing only 20 subjects -- which, however, is not unusual in the domain. \textbf{Future work} should extend the gender-based analysis and modelling to other types of voice recordings, \eg vowel productions and unscripted, variable content speech on larger databases.
\section{Acknowledgements}
This work was supported by MDSI -- Munich Data Science Institute as well as MCML -- Munich Center of Machine Learning. B.W.S is also with the Konrad Zuse School of Excellence in Reliable AI (relAI) in Munich, Germany.

\section{\refname}
\printbibliography[heading=none]

@article{Bryan14-GDI,
  title = {Gender Differences in the Association of Agitation and Suicide Attempts among Psychiatric Inpatients},
  author = {Bryan, Craig J. and Hitschfeld, Mario J. and Palmer, Brian A. and Schak, Kathryn M. and Roberge, Erika M. and Lineberry, Timothy W.},
  year = {2014},
  month = nov,
  journal = {Gen. Hosp. Psychiatry},
  volume = {36},
  number = {6},
  pages = {726--731},
%   issn = {01638343},
%   doi = {10.1016/j.genhosppsych.2014.09.013},
  langid = {american},
  keywords = {Agitation,Gender,Inpatient,Psychiatric hospitalization,Suicide}
}

@article{Stasak21-RSV,
  title = {Read Speech Voice Quality and Disfluency in Individuals with Recent Suicidal Ideation or Suicide Attempt},
  author = {Stasak, Brian and Epps, Julien and Schatten, Heather T. and Miller, Ivan W. and Provost, Emily Mower and Armey, Michael F.},
  year = {2021},
  month = sep,
  journal = {Speech Commun.},
  volume = {132},
  pages = {10--20},
%   issn = {01676393},
%   doi = {10.1016/j.specom.2021.05.004},
  langid = {american},
  keywords = {Digital medicine,Digital phenotyping,Machine learning,Mental health,Psychogenic voice disorders}
}

@article{Min23-AAO,
  title = {Acoustic Analysis of Speech for Screening for Suicide Risk: Machine Learning Classifiers for between- and within-Person Evaluation of Suicidality},
  shorttitle = {Acoustic Analysis of Speech for Screening for Suicide Risk},
  author = {Min, Sooyeon and Shin, Daun and Rhee, Sang Jin and Park, C Hyung Keun and Yang, Jeong Hun and Song, Yoojin and Kim, Min Ji and Kim, Kyungdo and Cho, Won Ik and Kwon, Oh Chul and Ahn, Yong Min and Lee, Hyunju},
  year = {2023},
  month = mar,
  journal = {J. Med. Internet Res.},
  volume = {25},
  number = {1},
  pages = {e45456},
  publisher = {{JMIR Publications Inc., Toronto, Canada}},
%   issn = {1438-8871},
%   doi = {10.2196/45456},
  langid = {american}
}

@article{Belouali21-AAL,
  title = {Acoustic and Language Analysis of Speech for Suicidal Ideation among {{US}} Veterans},
  author = {Belouali, Anas and Gupta, Samir and Sourirajan, Vaibhav and Yu, Jiawei and Allen, Nathaniel and Alaoui, Adil and Dutton, Mary Ann and Reinhard, Matthew J.},
  year = {2021},
  month = feb,
  journal = {BioData Min.},
  volume = {14},
  number = {1},
  pages = {11},
%   issn = {1756-0381},
%   doi = {10.1186/s13040-021-00245-y},
  langid = {american}
}

@article{Wagner23-DOT,
  title = {Dawn of the Transformer Era in Speech Emotion Recognition: Closing the Valence Gap},
  author = {Wagner, Johannes and Triantafyllopoulos, Andreas and Wierstorf, Hagen and Schmitt, Maximilian and Burkhardt, Felix and Eyben, Florian and Schuller, Bj{\"o}rn W.},
  year = {2023},
  month = sep,
  journal = {TPAMI},
  volume = {45},
  number = {9},
  pages = {10745--10759},
%   issn = {0162-8828, 2160-9292, 1939-3539},
%   doi = {10.1109/tpami.2023.3263585},
  langid = {american}
}

@article{Mann47-OAT,
  title = {On a Test of Whether One of Two Random Variables Is Stochastically Larger than the Other},
  author = {Mann, H. B. and Whitney, D. R.},
  year = {1947},
  journal = {Ann. Math. Stat.},
  volume = {18},
  number = {1},
  eprint = {2236101},
  eprinttype = {jstor},
  pages = {50--60},
  publisher = {{Institute of Mathematical Statistics}},
%   issn = {0003-4851},
%   doi = {10.1214/aoms/1177730491},
  langid = {american}
}

@article{Eyben16-TGM,
  title = {The Geneva Minimalistic Acoustic Parameter Set ({{GeMAPS}}) for Voice Research and Affective Computing},
  author = {Eyben, Florian and Scherer, Klaus R. and Schuller, Bjorn W. and Sundberg, Johan and Andre, Elisabeth and Busso, Carlos and Devillers, Laurence Y. and Epps, Julien and Laukka, Petri and Narayanan, Shrikanth S. and Truong, Khiet P.},
  year = {2016},
  month = apr,
  journal = {TAFFC},
  volume = {7},
  number = {2},
  pages = {190--202},
%   issn = {1949-3045},
%   doi = {10.1109/TAFFC.2015.2457417},
  langid = {american},
  keywords = {acoustic features,Acoustic Features,Affective computing,Affective Computing,emotion recognition,Emotion Recognition,Frequency measurement,geneva minimalistic parameter set,Geneva Minimalistic Parameter Set,Harmonic analysis,Licenses,Mel frequency cepstral coefficient,Speech,speech analysis,Speech Analysis,standard,Standard,Standards}
}

@inproceedings{Eyben10-OTM,
  title = {Opensmile: The Munich Versatile and Fast Open-Source Audio Feature Extractor},
  shorttitle = {Opensmile},
  booktitle = {ACM Multimedia 2010},
  author = {Eyben, Florian and W{\"o}llmer, Martin and Schuller, Bj{\"o}rn},
  year = {2010},
  month = oct,
  pages = {1459--1462},
  publisher = {{ACM Press}},
  address = {{Firenze, Italy}},
%   doi = {10.1145/1873951.1874246},
%   isbn = {978-1-60558-933-6},
  langid = {american}
}

@inproceedings{Baevski20-W2A,
  title = {Wav2vec 2.0: A Framework for Self-Supervised Learning of Speech Representations},
  shorttitle = {Wav2vec 2.0},
  booktitle = {Proc. NeurIPS 2020},
  author = {Baevski, Alexei and Zhou, Yuhao and Mohamed, Abdelrahman and Auli, Michael},
  year = {2020},
  volume = {33},
  address = {virtual},
  pages = {12449--12460},
  publisher = {{Curran Associates, Inc.}},
%   keywords = {No DOI found}
}

@article{Lotfian19-BNE,
  title = {Building Naturalistic Emotionally Balanced Speech Corpus by Retrieving Emotional Speech from Existing Podcast Recordings},
  author = {Lotfian, Reza and Busso, Carlos},
  year = {2019},
  month = oct,
  journal = {TAFFC},
  volume = {10},
  number = {4},
  pages = {471--483},
%   issn = {1949-3045, 2371-9850},
%   doi = {10.1109/TAFFC.2017.2736999},
  langid = {american}
}

@inproceedings{Honig14-AMO,
  title = {Automatic Modelling of Depressed Speech: Relevant Features and Relevance of Gender},
  shorttitle = {Automatic Modelling of Depressed Speech},
  booktitle = {Proc. INTERSPEECH 2014},
  author = {H{\"o}nig, Florian and Batliner, Anton and N{\"o}th, Elmar and Schnieder, Sebastian and Krajewski, Jarek},
  year = {2014},
  month = sep,
  pages = {1248--1252},
  publisher = {ISCA},
  address = {{Singapore}},
%   doi = {10.21437/Interspeech.2014-313},
  langid = {english}
}

@inproceedings{Cummins11-AIO,
  title = {An Investigation of Depressed Speech Detection: Features and Normalization},
  shorttitle = {An Investigation of Depressed Speech Detection},
  booktitle = {Proc. INTERSPEECH 2011},
  author = {Cummins, Nicholas and Epps, Julien and Breakspear, Michael and Goecke, Roland},
  year = {2011},
  month = aug,
  pages = {2997--3000},
  publisher = {ISCA},
  address = {{Firenze, Italy}},
%   doi = {10.21437/Interspeech.2011-750},
  langid = {american}
}

@inproceedings{Sethu09-SDO,
  title = {Speaker Dependency of Spectral Features and Speech Production Cues for Automatic Emotion Classification},
  booktitle = {{ICASSP} 2009},
  author = {Sethu, Vidhyasaharan and Ambikairajah, Eliathamby and Epps, Julien},
  year = {2009},
  month = apr,
  pages = {4693--4696},
  publisher = {IEEE},
  address = {Taipei, Taiwan},
%   doi = {10.1109/ICASSP.2009.4960678},
%   isbn = {978-1-4244-2353-8},
  langid = {english}
}

@article{Ozdas04-IOV,
  title = {Investigation of Vocal Jitter and Glottal Flow Spectrum as Possible Cues for Depression and Near-Term Suicidal Risk},
  author = {Ozdas, A. and Shiavi, R.G. and Silverman, S.E. and Silverman, M.K. and Wilkes, D.M.},
  year = {2004},
  month = sep,
  journal = {IEEE Trans. Biomed. Eng.},
  volume = {51},
  number = {9},
  pages = {1530--1540},
%   issn = {0018-9294},
%   doi = {10.1109/TBME.2004.827544},
  langid = {american}
}

@article{Guzman13-IOS,
  title = {Influence on Spectral Energy Distribution of Emotional Expression},
  author = {Guzman, Marco and Correa, Soledad and Mu{\~n}oz, Daniel and Mayerhoff, Ross},
  year = {2013},
  month = jan,
  journal = {J. Voice},
  volume = {27},
  number = {1},
  pages = {129.e1-129.e10},
%   issn = {08921997},
%   doi = {10.1016/j.jvoice.2012.08.008},
  langid = {american},
  keywords = {Actor,Emotions,LTAS,Spectral energy,Timbre,Voice quality}
}

@article{Waaramaa10-POE,
  title = {Perception of Emotional Valences and Activity Levels from Vowel Segments of Continuous Speech},
  author = {Waaramaa, Teija and Laukkanen, Anne-Maria and Airas, Matti and Alku, Paavo},
  year = {2010},
  month = jan,
  journal = {J. Voice},
  volume = {24},
  number = {1},
  pages = {30--38},
%   issn = {08921997},
%   doi = {10.1016/j.jvoice.2008.04.004},
  langid = {american},
  keywords = {Inverse filtering,Perception of emotional valence,Voice quality}
}

@article{Hammarberg80-PAA,
  title = {Perceptual and Acoustic Correlates of Abnormal Voice Qualities},
  author = {Hammarberg, Britta and Fritzell, B. and Gaufin, J. and Sundberg, J. and Wedin, L.},
  year = {1980},
  month = jan,
  journal = {Acta Otolaryngol.},
  volume = {90},
  number = {1-6},
  pages = {441--451},
%   issn = {0001-6489, 1651-2251},
%   doi = {10.3109/00016488009131746},
  langid = {american},
  pmid = {7211336},
  keywords = {Humans,Speech Acoustics,Speech Perception,Voice,Voice Disorders,Voice Quality,Voice Training}
}

@article{Banse96-API,
  title = {Acoustic Profiles in Vocal Emotion Expression.},
  author = {Banse, Rainer and Scherer, Klaus R.},
  year = {1996},
  journal = {J. Pers. Soc. Psychol.},
  volume = {70},
  number = {3},
  pages = {614--636},
%   issn = {1939-1315, 0022-3514},
%   doi = {10.1037/0022-3514.70.3.614},
  langid = {american}
}

@article{Cummins15-ARO,
  title = {A Review of Depression and Suicide Risk Assessment Using Speech Analysis},
  author = {Cummins, Nicholas and Scherer, Stefan and Krajewski, Jarek and Schnieder, Sebastian and Epps, Julien and Quatieri, Thomas F.},
  year = {2015},
  month = jul,
  journal = {Speech. Commun.},
  volume = {71},
  pages = {10--49},
%   issn = {01676393},
%   doi = {10.1016/j.specom.2015.03.004},
  langid = {american}
}

@inproceedings{Vlasenko17-IGV,
  title = {Implementing Gender-Dependent Vowel-Level Analysis for Boosting Speech-Based Depression Recognition},
  booktitle = {Proc. INTERSPEECH 2017},
  author = {Vlasenko, Bogdan and Sagha, Hesam and Cummins, Nicholas and Schuller, Bj{\"o}rn},
  year = {2017},
  month = aug,
  pages = {3266--3270},
  publisher = {ISCA},
  address = {Stockholm, Sweden},
%   doi = {10.21437/Interspeech.2017-887},
  langid = {american}
}

@book{Cummins22-MMO,
  title = {Multilingual Markers of Depression in Remotely Collected Speech Samples},
  author = {Cummins, Nicholas and Dineley, Judith and Conde, Pauline and Matcham, Faith and Siddi, Sara and Lamers, Femke and Carr, Ewan and Lavelle, Grace and Leightley, Daniel and White, Katie and Oetzmann, Carolin and Campbell, Edward and Simblett, Sara and Bruce, Stuart and Haro, Josep Maria and Penninx, Brenda and Ranjan, Yatharth and Rashid, Zulqarnain and Stewart, Callum and Hotopf, Matthew},
  year = {2022},
  month = oct,
%   doi = {10.21203/rs.3.rs-2183980/v1}
}

@inproceedings{Bain23-WTS,
  title = {{{WhisperX}}: Time-Accurate Speech Transcription of Long-Form Audio},
  shorttitle = {{{WhisperX}}},
  booktitle = {Proc. INTERSPEECH 2023},
  author = {Bain, Max and Huh, Jaesung and Han, Tengda and Zisserman, Andrew},
  year = {2023},
  month = aug,
  address = {Dublin, Ireland},
  pages = {4489--4493},
%   doi = {10.21437/Interspeech.2023-78},
  langid = {american}
}

@article{Iyer22-DOS,
  title = {Detection of Suicide Risk Using Vocal Characteristics: Systematic Review},
  shorttitle = {Detection of Suicide Risk Using Vocal Characteristics},
  author = {Iyer, Ravi and Meyer, Denny},
  year = {2022},
  month = dec,
  journal = {JMIR Biomed. Eng.},
  volume = {7},
  number = {2},
  pages = {e42386},
  publisher = {JMIR Publications Inc., Toronto, Canada},
%   issn = {2561-3278},
%   doi = {10.2196/42386},
  copyright = {Unless stated otherwise, all articles are open-access distributed under the terms of the Creative Commons Attribution License (http://creativecommons.org/licenses/by/2.0/), which permits unrestricted use, distribution, and reproduction in any medium, provided the original work ("first published in the Journal of Medical Internet Research...") is properly cited with original URL and bibliographic citation information. The complete bibliographic information, a link to the original publication on http://www.jmir.org/, as well as this copyright and license information must be included.},
  langid = {american}
}

@article{Sudol17-BOS,
  title = {Biomarkers of Suicide Attempt Behavior: Towards a Biological Model of Risk},
  shorttitle = {Biomarkers of Suicide Attempt Behavior},
  author = {Sudol, Katherin and Mann, J. John},
  year = {2017},
  month = jun,
  journal = {Curr. Psychiatry Rep.},
  volume = {19},
  number = {6},
  pages = {31},
%   issn = {1523-3812, 1535-1645},
%   doi = {10.1007/s11920-017-0781-y},
  langid = {american},
  keywords = {Biomarkers,Brain imaging,Genetics,Suicidal behavior,Suicide attempt,Suicide risk}
}

@inproceedings{Scherer13-ITS,
  title = {Investigating the Speech Characteristics of Suicidal Adolescents},
  booktitle = {{ICASSP}},
  author = {Scherer, Stefan and Pestian, John and Morency, Louis-Philippe},
  year = {2013},
  month = may,
  pages = {709--713},
  address = {Vancouver, Canada},
%   issn = {2379-190X},
%   doi = {10.1109/ICASSP.2013.6637740},
  langid = {american},
  keywords = {Accuracy,Acoustic measurements,Acoustics,classification,Feature extraction,Hidden Markov models,Interviews,Speech,speech characteristics,Suicide prevention,voice quality,voice source model}
}

@article{Franklin17-RFF,
  title = {Risk Factors for Suicidal Thoughts and Behaviors: A Meta-Analysis of 50 Years of Research.},
  shorttitle = {Risk Factors for Suicidal Thoughts and Behaviors},
  author = {Franklin, Joseph C. and Ribeiro, Jessica D. and Fox, Kathryn R. and Bentley, Kate H. and Kleiman, Evan M. and Huang, Xieyining and Musacchio, Katherine M. and Jaroszewski, Adam C. and Chang, Bernard P. and Nock, Matthew K.},
  year = {2017},
  journal = {Psychol. Bull.},
  volume = {143},
  number = {2},
  pages = {187--232},
%   issn = {1939-1455, 0033-2909},
%   doi = {10.1037/bul0000084},
  langid = {american}
}

@article{Cummins18-SAF,
  title = {Speech Analysis for Health: Current State-of-the-Art and the Increasing Impact of Deep Learning},
  author = {Cummins, Nicholas and Baird, Alice and Schuller, Bj{\"o}rn W.},
  year = {2018},
  month = dec,
  journal = {Methods},
  volume = {151},
  pages = {41--54},
  publisher = {Elsevier},
%   issn = {10462023},
%   doi = {10.1016/j.ymeth.2018.07.007},
  langid = {american}
}

@article{Yan23-SBA,
  title = {Suicide before and during the {{COVID-19}} Pandemic: A Systematic Review with Meta-Analysis},
  shorttitle = {Suicide before and during the {{COVID-19}} Pandemic},
  author = {Yan, Yifei and Hou, Jianhua and Li, Qing and Yu, Nancy Xiaonan},
  year = {2023},
  month = feb,
  journal = {Int. J. Environ. Res. Public. Health},
  volume = {20},
  number = {4},
  pages = {3346},
%   issn = {1660-4601},
%   doi = {10.3390/ijerph20043346},
  langid = {american},
  pmcid = {PMC9960664},
  pmid = {36834037}
}

@inproceedings{Oureshi21-GEO,
  title = {Gender-Aware Estimation of Depression Severity Level in a Multimodal Setting},
  booktitle = {{IJCNN}},
  author = {Oureshi, Syed Arbaaz and Dias, Gael and Saha, Sriparna and Hasanuzzaman, Mohammed},
  year = {2021},
  month = jul,
  pages = {1--8},
  address = {virtual},
%   issn = {2161-4407},
%   doi = {10.1109/IJCNN52387.2021.9534330},
  langid = {american},
  keywords = {Benchmark testing,Depression,Estimation,Gender,Interviews,Mood,Multitask learning,Neural networks,Task analysis}
}

@article{world2021suicide,
  title={Suicide worldwide in 2019: global health estimates},
  author={{Geneva: World Health Organization}},
  year={2021},
  publisher={World Health Organization}
}

@article{Canetto98-TGP,
  title = {The Gender Paradox in Suicide},
  author = {Canetto, Silvia Sara and Sakinofsky, Isaac},
  year = {1998},
  journal = {Suicide Life. Threat. Behav.},
  volume = {28},
  number = {1},
  pages = {1--23},
%   issn = {1943-278X},
%   doi = {10.1111/j.1943-278X.1998.tb00622.x},
  langid = {english}
}

@article{Callanan12-GDI,
  title = {Gender Differences in Suicide Methods},
  author = {Callanan, Valerie J. and Davis, Mark S.},
  year = {2012},
  month = jun,
  journal = {Soc. Psychiatry Psychiatr. Epidemiol.},
  volume = {47},
  number = {6},
  pages = {857--869},
%   issn = {0933-7954, 1433-9285},
%   doi = {10.1007/s00127-011-0393-5},
  langid = {american},
  keywords = {Firearms,Gender,Hanging,Poisoning,Suicide methods}
}

@article{Ludwig18-TCO,
  title = {The Concept of Violent Suicide, Its Underlying Trait and Neurobiology: {{A}} Critical Perspective},
  shorttitle = {The Concept of Violent Suicide, Its Underlying Trait and Neurobiology},
  author = {Ludwig, Birgit and Dwivedi, Yogesh},
  year = {2018},
  month = feb,
  journal = {Eur. Neuropsychopharmacol.},
  volume = {28},
  number = {2},
  pages = {243--251},
%   issn = {0924977X},
%   doi = {10.1016/j.euroneuro.2017.12.001},
  langid = {american},
  keywords = {Aggression,Neurobiology,Self-directed violence,Suicide,Violence,Violent suicide}
}

@article{Cibis12-POL,
  title = {Preference of Lethal Methods Is Not the Only Cause for Higher Suicide Rates in Males},
  author = {Cibis, Anna and Mergl, Roland and Bramesfeld, Anke and Althaus, David and Niklewski, G{\"u}nter and Schmidtke, Armin and Hegerl, Ulrich},
  year = {2012},
  month = jan,
  journal = {J. Affect. Disord.},
  volume = {136},
  number = {1-2},
  pages = {9--16},
%   issn = {01650327},
%   doi = {10.1016/j.jad.2011.08.032},
  langid = {american},
  keywords = {Gender,Lethality,Methods,Suicide,Suicide prevention}
}

@article{Freeman17-ACS,
  title = {A Cross-National Study on Gender Differences in Suicide Intent},
  author = {Freeman, Aislinn{\'e} and Mergl, Roland and Kohls, Elisabeth and Sz{\'e}kely, Andr{\'a}s and Gusmao, Ricardo and Arensman, Ella and Koburger, Nicole and Hegerl, Ulrich and {Rummel-Kluge}, Christine},
  year = {2017},
  month = dec,
  journal = {BMC Psychiatry}, 
  volume = {17},
  number = {1},
  pages = {234},
%   issn = {1471-244X},
%   doi = {10.1186/s12888-017-1398-8},
  langid = {english},
  keywords = {Attempt,Gender differences,Intent,Suicidal behaviour,Suicide}
}

@article{Milner20-SIG,
  title = {Shifts in Gender Equality and Suicide: {{A}} Panel Study of Changes over Time in 87 Countries},
  shorttitle = {Shifts in Gender Equality and Suicide},
  author = {Milner, Allison and Scovelle, Anna J. and Hewitt, Belinda and Maheen, Humaira and Ruppanner, Leah and King, Tania L.},
  year = {2020},
  month = nov,
  journal = {J. Affect. Disord.},
  volume = {276},
  pages = {495--500},
%   issn = {01650327},
%   doi = {10.1016/j.jad.2020.07.105},
  langid = {american},
  keywords = {Fixed effects,Gender,Gender equality,Panel study,Suicide}
}

\end{document}